\documentclass{article}

\usepackage[preprint]{neurips_2025}

\usepackage[utf8]{inputenc} 
\usepackage[T1]{fontenc}    
\usepackage{hyperref}       
\usepackage{url}            
\usepackage{booktabs}       
\usepackage{amsfonts}       
\usepackage{nicefrac}       
\usepackage{microtype}      
\usepackage{xcolor}         
\usepackage{multirow}
\usepackage{graphicx}
\usepackage{amsmath}  
\usepackage{wrapfig}
\usepackage{subcaption}
\usepackage{adjustbox}

\definecolor{earthyellow}{rgb}{0.88, 0.66, 0.37}
\definecolor{carrotorange}{rgb}{0.93, 0.57, 0.13}
\definecolor{jiangweicolor}{rgb}{0.8, 0.36, 0.36}
\definecolor{gaojiecolor}{rgb}{0.47, 0.53, 0.6}
\definecolor{liangchaocolor}{rgb}{0.13, 0.54, 0.13}

\title{AlignHuman: Improving Motion and Fidelity via Timestep-Segment Preference Optimization for Audio-Driven Human Animation}

\author{
\textbf{Chao Liang$^{*}$},~ 
Jianwen Jiang$^{* \dag}$,~
Wang Liao$^{*}$,~ 
\textbf{Jiaqi Yang},~
\textbf{Zerong Zheng},~
\and
\textbf{Weihong Zeng},~
\textbf{Han Liang}
\\
ByteDance\\
{\footnotesize \texttt{\{liangchao.0412,jianwen.alan,lw46756583\}@gmail.com}}\\
\textsuperscript{${*}$}Equal contribution, 
\textsuperscript{\dag}Project lead
}

\begin{document}
\maketitle
\begin{abstract}
Recent advancements in human video generation and animation tasks, driven by diffusion models, have achieved significant progress. However, expressive and realistic human animation remains challenging due to the trade-off between motion naturalness and visual fidelity. To address this, we propose \textbf{AlignHuman}, a framework that combines Preference Optimization as a post-training technique with a divide-and-conquer training strategy to jointly optimize these competing objectives.  
Our key insight stems from an analysis of the denoising process across timesteps: (1) early denoising timesteps primarily control motion dynamics, while (2) fidelity and human structure can be effectively managed by later timesteps, even if early steps are skipped. Building on this observation, we propose timestep-segment preference optimization (TPO) and introduce two specialized LoRAs as expert alignment modules, each targeting a specific dimension in its corresponding timestep interval. The LoRAs are trained using their respective preference data and activated in the corresponding intervals during inference to enhance motion naturalness and fidelity.
Extensive experiments demonstrate that AlignHuman improves strong baselines and reduces NFEs during inference, achieving a 3.3$\times$ speedup (from 100 NFEs to 30 NFEs) with minimal impact on generation quality. Homepage: \href{https://alignhuman.github.io/}{https://alignhuman.github.io/}

\end{abstract}

\section{Introduction}

Audio-driven Human Animation aims to synthesize dynamic human videos conditioned on a reference image and driving audio. Recent works~\cite{lincyberhost,zhuang2024vlogger,tian2025emo2,lin2025omnihuman,yi2025magicinfinite,wei2025mocha,wang2025fantasytalking,meng2024echomimicv2,qiu2025skyreels} have made notable progress in facial expressions, motion diversity, and visual quality. However, as the overall performance improves, certain challenges become more pronounced and harder to address, such as unnatural movements, reduced realism, and uncanny valley effects. 
Preference optimization~\cite{rafailov2023direct,azar2024general, meng2024simpo, wallace2024diffusion} has proven effective to conduct preference alignment as a post-training scheme in both large language models (LLMs)~\cite{grattafiori2024llama, yang2025qwen2, mehta2024openelm} and video generation~\cite{seawead2025seaweed, chen2025skyreels, he2024videoscore} tasks. By incorporating preference optimization, models are expected to improve based on subjective human feedback. However, directly applying it to human animation does not lead to significant improvements.


The multi-objective task often overfits on simpler objectives or biases the model toward learning the most prominent preferences. Existing multi-dimensional preference optimization methods~\cite{chen2025skyreels, xu2024visionreward, chen2025skyreels} address this by enforcing strict rules during preference dataset collection to ensure sufficient diversity between paired samples.
We find that such strategies fall short for domain-specific human animation tasks. For motion naturalness, preference data collection is often influenced by subjective biases. A sample with better motion might feature large or small movements, or ambiguous qualities lacking a consistent standard. Conversely, samples with worse motion frequently exhibit shared issues, such as poor synthesis quality, which are also heavily marked in fidelity-related preference data. During alignment training, this can cause the model to focus on fixing common issues in "bad" samples, such as poor local synthesis or low fidelity, while neglecting the motion-specific problems the motion-related preference pairs were intended to address.

\begin{figure}[tbp]
    \centering
    \includegraphics[width=\linewidth]{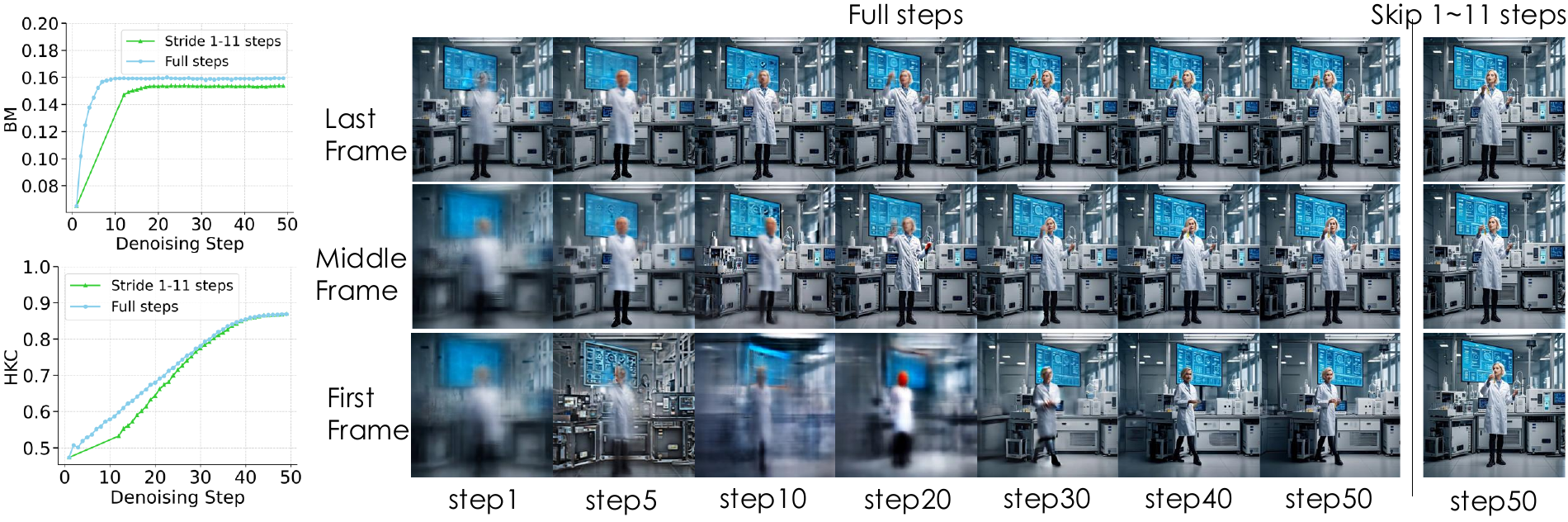}
        \caption{The analysis of intermediate performance at different denoising timesteps. \textbf{Left}: Temporal evolution of fidelity and motion across timesteps. \textbf{Right:} The visualization result. It's obvious that motion dynamics is primarily established during early denoising stages, while fidelity improves progressively throughout the whole process. By step-skipping the motion interval, the generated videos exhibit significant motion variation while maintaining nearly identical fidelity.}
        \label{timestep:vis}
        \vspace{-2pt}
\end{figure}

To investigate the joint optimization of motion and fidelity trade-offs, we first analyze the denoising timesteps of a trained diffusion model using two metrics (BM, Body Motion; HKC, Hand Keypoint Confidence) to separately evaluate motion and human fidelity. We also report the visualization results. As shown in Fig.~\ref{timestep:vis}, human motion is largely determined during the early denoising stages, while later steps have little influence on motion dynamics. In contrast, human fidelity improves steadily throughout the process, with skipping early timesteps having little impact on the final quality.
This insight motivates us to leverage the biases in fidelity and motion generation across timesteps, along with their corresponding preference data, to develop a divide-and-conquer method for optimizing both fidelity and motion in the preference learning process, named AlignHuman.

Based on the above observations and analysis, AlignHuman introduces two key designs: First, a timestep-segment preference optimization strategy that decouples the Direct Preference Optimization (DPO) training into distinct phases. During early denoising timesteps, motion naturalness preference pairs are prioritized to capture fluid dynamics, while synthesis fidelity preference pairs are leveraged in late timesteps to refine fine-grained details.
Second, specialized LoRA preference experts activated by timestep segments: A motion LoRA module is dynamically engaged during early stages to model temporal coherence, and  a fidelity LoRA would be activated in late stages to enhance structural fidelity. This timestep-dependent activation scheme is rigorously preserved during inference. This simple yet effective design allows the model to optimize both motion and synthesis fidelity more specifically, avoiding the common issue of over-prioritizing fidelity at the expense of motion. Experimental results demonstrate that AlignHuman achieves significant improvements over current state-of-the-art models by balancing motion naturalness and visual fidelity.

Our contributions are summarized as: (1) We investigate the denoising diffusion model in human animation, revealing distinct bias for motion naturalness and fidelity across denoising timesteps. (2) Based on this insight, we propose a timestep-segment preference omtimization (TPO) method with specialized LoRA modules to disentangle motion and fidelity alignment learning. (3) Our proposed method achieves significant performance, gaining state-of-the-art audio-driven experiment results. Moreover, it reduces inference NFEs, delivering a 3.3$\times$ speed-up with acceptable generation quality.

\section{Related Works}

\noindent{\bf Human Animation} generates human videos using input images and driving signals such as audio or video sequences. Early GAN-based~\cite{siarohin2019first,wang2018video, siarohin2021motion} and the more advanced Diffusion-based~\cite{lincyberhost,wang2025fantasytalking,meng2024echomimicv2,lin2025omnihuman,wei2025mocha,zhang2024mimicmotion,jiang2024loopy, yi2025magicinfinite} works represent two key technique paradigms.
In terms of the audio-driven setting, existing works~\cite{jiang2024loopy,tian2024emo,chen2025echomimic} initially focus on portrait animation, where only head regions could be driven. Recently, some works~\cite{lincyberhost, yi2025magicinfinite, wang2025fantasytalking, meng2024echomimicv2, tian2025emo2} seek to generate videos with different character types and styles, including half-body, full-body, or cartoons. VLogger~\cite{zhuang2024vlogger} first supports zero-shot generation without relying on face cropping. To improve hand quality, CyberHost~\cite{lincyberhost} designs a codebook module to provide more local visual features, and EchomimicV2~\cite{meng2024echomimicv2} employs pose condition as an auxiliary
driving signal. OmniHuman-1~\cite{lin2025omnihuman} investigates the scaling effects in data by leveraging weaker text-to-video conditions to scale up audio-driven task, and accommodates different image styles. MagicInfinity~\cite{yi2025magicinfinite}
employs a two-stage curriculum learning scheme to integrate multi-conditions for different objectives and proposes a CFG distillation to achieve inference speedup.

\noindent{\bf Preference Learning} aims to align model outputs with human preferences and has been widely applied in  language~\cite{grattafiori2024llama, yang2025qwen2, mehta2024openelm,zhou2020learning,stiennon2020learning,rafailov2023direct}, vision~\cite{wallace2024diffusion, xu2023imagereward, chen2025skyreels,liu2025improving,he2024videoscore,li2024t2v,liu2024videodpo,xu2024visionreward} and multi-modal~\cite{ouali2024clip, tong2024cambrian} tasks. In this field, reward-based~\cite{xu2023imagereward,wu2023human,li2024t2v,he2024videoscore,xu2024visionreward} and data-based~\cite{wallace2024diffusion,meng2024simpo,azar2024general,rafailov2023direct,liu2024videodpo} methods are commonly used approaches. The reward-based method optimizes the policy model by maximizing explicit reward signals, which could be from the environment or the specific reward models. And the data-based method directly maximizes the likelihood of preference data. The key difference among data-based methods lies in their various preference-optimization loss functions. DPO~\cite{rafailov2023direct} optimizes the log probability ratio between preferred and dispreferred responses while constraining deviation from the original distribution via a reference model. To avoid over-optimization, IPO~\cite{azar2024general} constrains the log probability difference between preferred and dispreferred responses to approximate a hyperparameter representing the preference intensity. SimPO\cite{meng2024simpo} proposes to employ the average logarithmic probability of the sequence as the implicit reward to better align the generation process. Our work proposes a data-based method to optimize competing dimensions in human animation.

\section{Method}
In this section, we present the detailed design of AlignHuman. In Sec.~\ref{sec:basemodel}, we introduce a diffusion-based base model of the audio-driven human animation. Sec.~\ref{sec:analysis} presents an analysis of how different denoising timesteps affect motion generation and visual fidelity of the intermediate videos. After that, Sec.~\ref{sec:tpo} outlines our timestep-segment preference optimization pipeline, including: (1) the protocol for constructing the dimension-specific preference data, (2) the divide-and-conquer training strategy and the dimension-specific expert LoRA modules for preference optimization, and (3) the implementation details for training and inference.

\subsection{Base Model}
\label{sec:basemodel}
We first introduce our human animation base model. It is based on the pretrained text-to-video MMDiT model ~\cite{seawead2025seaweed,peebles2023scalable}, comprising a 3D Variational Autoencoder (VAE)~\cite{kingma2013auto} and a latent Diffusion Transformer (DiT)~\cite{peebles2023scalable}. To adapt the base text-to-video model for audio-driven human animation, we implement architectural modifications following the current state-of-the-art method~\cite{lin2025omnihuman}. For audio signals, we extract audio features using Wav2Vec2.0~\cite{baevski2020wav2vec} and align their dimensions via a lightweight MLP, then inject them into each MMDiT block through cross-attention layers. For the reference image, we downsample it using the VAE encoder and then concatenate it with the noise latents. The interaction between the reference image and the video latents is mediated through self-attention layers. Taking the reference image, driving audio, and text prompt as inputs, the base model $\theta$ is trained using a flow matching optimization objective~\cite{lipman2022flow}. The loss function is defined as

\begin{equation}
\mathcal{L} = \mathbb{E}_{t, z_0, \epsilon} \left\| v_{\theta}(z_t, t, c) - (z_1 - z_0) \right\|_2^2
\label{eq:loss_function}
\end{equation}

where $c=(c_{\text{text}}, c_{\text{audio}}, c_{\text{reference}})$ represent the conditions and $(z_1 - z_0)$ is the ground truth field velocity.

\subsection{Timestep Impact on Motion and Fidelity}
\label{sec:analysis}
We first attempted to directly apply the Direct Preference Optimization (DPO) method~\cite{rafailov2023direct,wallace2024diffusion} for preference learning. However, as shown in Tab.~\ref{tab:ablation_key_design} in the experiment section, this approach was not effective.
Although we have ensured that the win samples in the preference data outperform the lose samples in both motion and fidelity dimensions,
the disparity in learning difficulty between the two objectives still introduces ambiguity to the training. As a result, the model prioritized learning the easier fidelity objectives to resolve synthesis artifact issues, while motion naturalness, the higher-level semantic preferences, remained challenging to capture, resulting in limited improvements.

Prior works on noise scheduling~\cite{karras2022elucidating, lu2022dpm} suggest that certain denoising steps play critical roles, motivating us to explore the impact of different timesteps in our base model. To validate this, we conduct inference using a small batch of the validation set, decoding the denoised latents at each timestep into final video frames. Since the training objective is based on flow matching, the final denoised results at intermediate steps can be obtained by leveraging their current velocity. The results are evaluated using two metrics, BM (Body Motion) and HKC (Hand Keypoint Confidence), to measure motion dynamics and visual fidelity. In current video generation models, hands are a key indicator of human fidelity and a frequent source of synthesis issues. Poor hand generation often leads to lower scores in hand keypoint detection. The findings are shown in Fig.~\ref{timestep:vis}.

As shown by the blue curve in Fig.~\ref{timestep:vis}, motion dynamics are primarily established during the early timesteps, with the BM metric stabilizing after just 10 denoising steps. To further investigate, we added a green curve representing results where the first 10 steps (the motion becomes stable here) were skipped, revealing noticeable motion differences between the final outputs and those obtained through complete denoising. In contrast, the visual fidelity metric improves gradually throughout the entire denoising process. Skipping the first 10 steps has little impact on the final visual fidelity. This suggests that the denoising timesteps play distinct roles, with motion generation relying on early intervals and visual fidelity benefiting from the late process. The intermediate timestep visualizations provided  further support for this finding.

\begin{figure}[tbp]
    \centering
    \includegraphics[width=\linewidth]{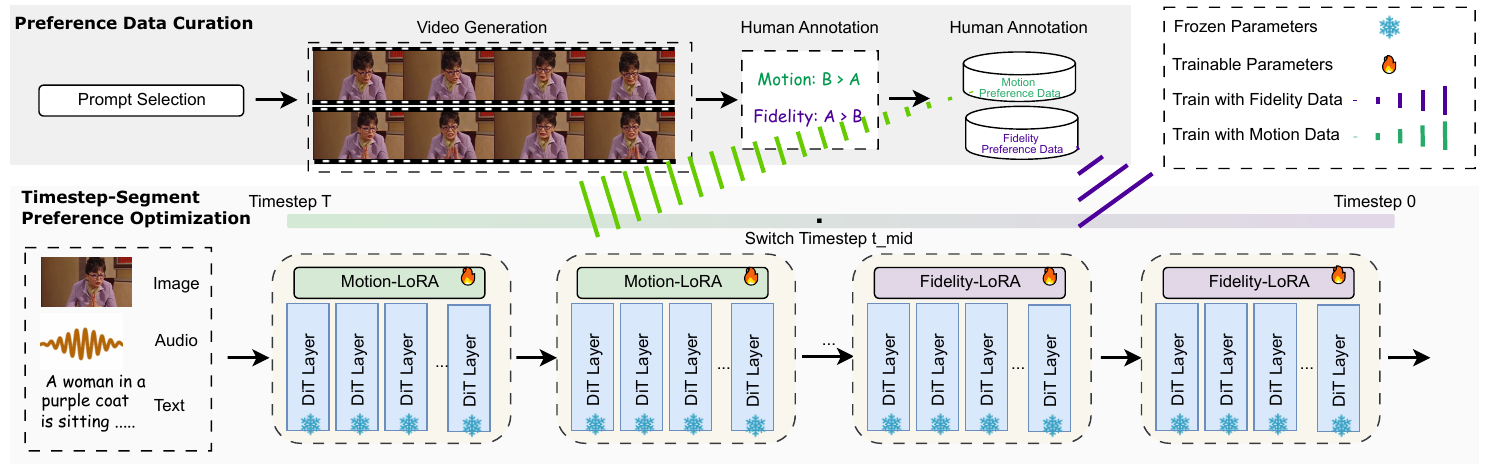}
        \caption{The framework of our timestep-segment preference optimization.} 
        \label{fig:framework}
\end{figure}
\subsection{Performance Optimization}
Building on the above discussion, this subsection introduces the construction of our preference learning dataset, followed by the proposed timestep-segment preference optimization (TPO) method and the design of LoRA modules to specialize the learning of motion and video fidelity within their respective timestep intervals, as shown in Fig.~\ref{fig:framework}.
\label{sec:tpo}
\subsubsection{Preference Dataset Curation}
\label{sec:dataset_curation}
The preference dataset is constructed through three stages: prompt (input images and audios) selection, video generation, and human annotation. To ensure the diversity and balance, we leverage a VLM tool~\cite{wang2024qwen2} and an MLLM tool~\cite{yao2024minicpm} to curate the prompts from OpenHumanVid~\cite{li2024openhumanvid}, and the selected prompts cover a range of dimensions of image styles, character types, audio languages, etc. For video generation, we trained nine candidate models with varying training recipes (randomly divided dataset subsets and different CFGs). For each curated prompt, four candidate models were randomly selected for every prompt to generate the video outputs. This process resulted in 10,000 video pairs, each with four videos, totaling 40,000 samples for human annotation. During human annotation, the annotators rank the four videos within a pair separately for visual fidelity and motion quality. This means one video may rank higher in fidelity but lower in motion quality. Severely poor-quality negative samples are flagged and discarded to ensure the preference dataset remains matched to the base model's distribution.

\subsubsection{Timestep-Segment Preference Optimization}

\textbf{Timestep Segment Strategy. }Building on the timestep analysis in Sec.~\ref{sec:analysis}, we introduce a simple yet effective solution, Timestep-Segment Preference Optimization (TPO), which separates the motion and fidelity preference optimization into their respective critical timestep intervals. Unlike conventional flow matching training, which uniformly samples timesteps across [0, T], TPO introduces a switch timestep \( t_{\text{mid}} \) to divide the timestep range. The early denoising interval [$T$ - \( t_{\text{mid}} \), $T$) is allocated for motion dimension, while the later interval [0, $T$ - \( t_{\text{mid}} \)] is dedicated to fidelity dimension.
To ensure balanced learning, timesteps are sampled with equal probability from both intervals during training, then the corresponding preference data is used to optimize the respective dimension.

\textbf{Specialized LoRA Design.} While dividing timestep intervals is expected to help the preference learning simultaneously optimize both the motion and fidelity, we found that fully fine-tuning the entire model with the limited preference data does not yield the best results. Therefore, we propose training in a lightweight manner: keeping the base model’s parameters fixed and introducing two expert LoRA modules—Motion LoRA and Fidelity LoRA. The two LoRAs are applied to all linear layers of all DiT blocks. During training and inference, each expert LoRA is activated only within the timestep interval corresponding to its designated dimension. Since the two timestep intervals do not overlap, only one expert module is active at any given denoising timestep. This design ensures effective optimization of motion and fidelity preferences in their critical intervals while maintaining stability and reliable performance.

\textbf{Training Pipeline.} During training, we first initialize the policy model $\theta$ and the frozen reference model $\text{ref}$, using weights from the trained base model. We then attach LoRA expert modules to the policy model and freeze all parameters except those of the LoRA modules. The pair of win sample $x^w_d$ and lose sample $x^l_d$ in dimesnsion $d$ are fed into both the policy and reference models to compute the original flow matching loss using Equ.~\ref{eq:loss_function}. The final loss function for preference optimization is defined as follows, where $\sigma$ is the sigmoid function and  $\beta=50$ here is a hyperparameter.

\begin{equation}
\mathcal{L}_d = -log\sigma(-\frac{\beta}{2}[(\mathcal{L}_{\mathrm{\theta}}^{w_d} - \mathcal{L}_{\theta}^{l_d})-(\mathcal{L}_{\mathrm{ref}}^{w_d} - \mathcal{L}_{\mathrm{ref}}^{l_d})])
\label{eq:TPO_loss_function}
\end{equation}

\section{Experiments}
\subsection{Experimental Settings}
 
\noindent \textbf{Dataset.} We train the base model using 2,000 hours of in-house audio-visual data and 8,000 hours of in-house text-to-video data, 
ensuring quality through filtering techniques including clarity scoring, cut detection, aesthetic scoring, OCR detection, lip-sync detection, etc. 
For TPO training, we curate a preference dataset of 10,000 video pairs, where each pair includes four videos ranked by human annotators for motion naturalness and visual fidelity, as described in Sec~\ref{sec:dataset_curation}.  Evaluation is conducted on a benchmark test set following prior work~\cite{lincyberhost}, featuring 269 samples that cover a wide range of scenes, genders, initial poses, and audio content.

\noindent \textbf{Implementation Details.} We train the base model at the batch size of 64 over two stages, with each stage running for 10 days. In the first stage, the model is trained for text-to-video generation using text-related data. In the second stage, audio data is introduced, maintaining a 50\% sampling ratio between audio and text data, to train the model for audio-driven human animation. Next, we perform preference learning using the preference dataset with a batch size of 8 for two epochs. At each step, there is a 50\% probability to optimize either the motion or fidelity dimension. Both base and TPO models are trained with the AdamW optimizer, using learning rates of 5e-5 and 1e-6, respectively.
\begin{figure}[tbp]
    \centering
    \includegraphics[width=\linewidth]{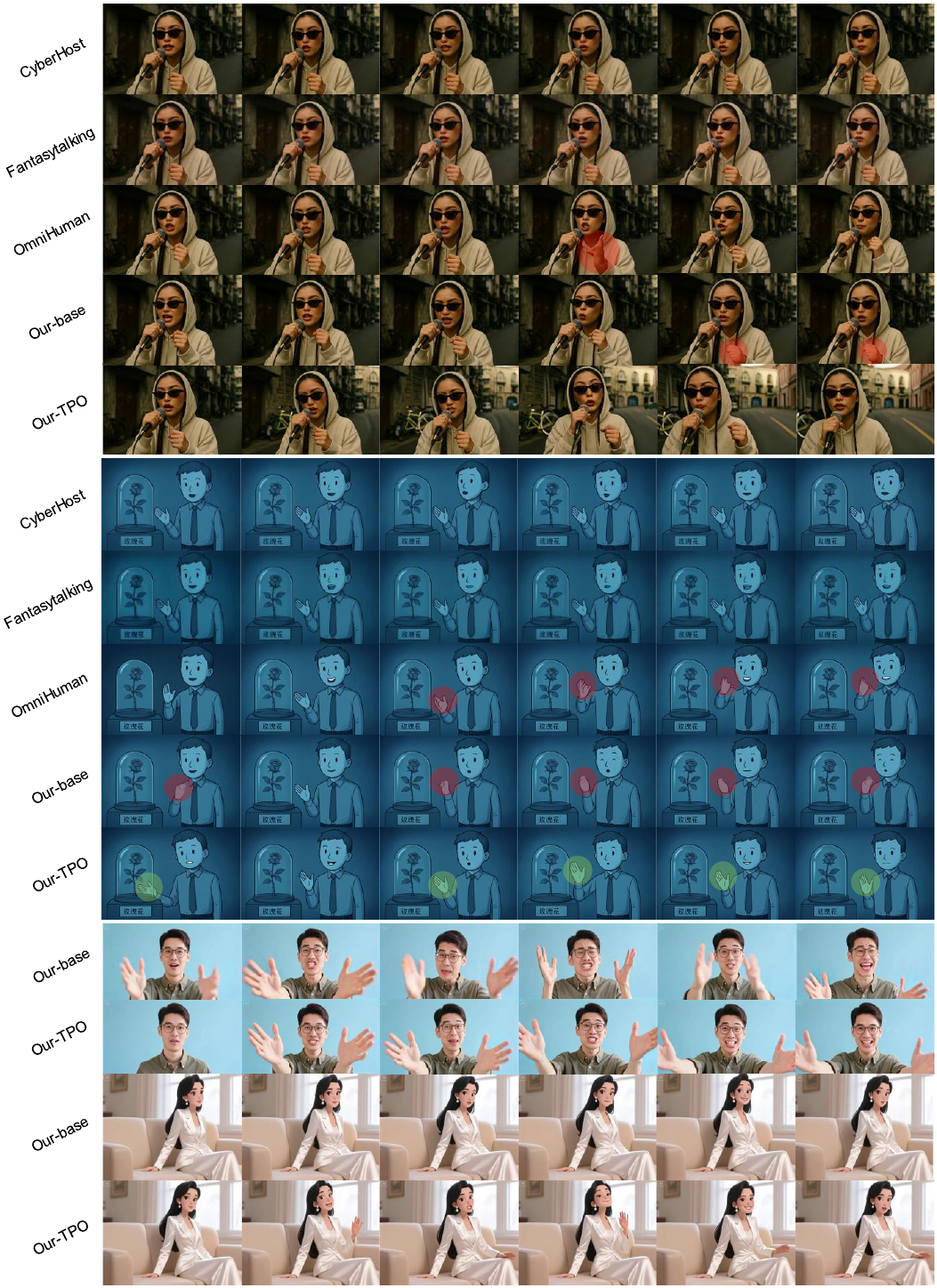}
    \caption{Visualization results. Compared with baselines and the base model, our TPO model can generate natural and rich character, background, and camera movements. Furthermore, our TPO method helps generate high-fidelity details of human hands and faces,  and it also completes the reasonable background when the camera moves.}
    
    \label{fig:vis}
\end{figure}

\noindent \textbf{Evaluation Baselines and Metrics}
We compare our approach with recent end-to-end human animation methods, including DiffTED~\cite{hogue2024diffted}, CyberHost~\cite{lincyberhost}, OmniHuman-1~\cite{lin2025omnihuman}, and FantasyTalking~\cite{wang2025fantasytalking}. Additionally, we evaluate against the two-stage pipeline DiffGest~\cite{zhu2023taming} + Mimicmotion~\cite{zhang2024mimicmotion}, which first predicts gesture skeletons from audio and then generates the final video using a video-to-video rendering process.
For metrics, we use Frechet Inception Distance (FID)~\cite{heusel2017gans} and Fréchet Video Distance (FVD)~\cite{unterthiner2019fvd} to assess overall generation quality. To evaluate specific aspects of visual quality (IQA) and aesthetics (AES), we apply q-align~\cite{wu2023q} to score the generated videos. Sync-C~\cite{chung2017out} is used to measure the confidence of lip-sync. Finally, we employ HKC and HKV~\cite{lincyberhost} to evaluate hand quality and motion richness in the generated videos.

\subsection{Comparisons with State-of-the-Arts Methods}

\noindent \textbf{Qualitative Results.} Fig.~\ref{fig:vis} presents a comprehensive temporal comparison of videos (left to right) across all methods. (1) Top: The TPO model exhibits a significant advantage in generating natural and dynamic motions, with rich and varied camera movements and accurate lip-sync. In contrast, primarily generating lip movements while the rest of the elements remain static or repetitive. (2) Middle: The TPO model demonstrates robust preservation of visual fidelity, particularly in generating intricate local features (e.g., facial attributes, textual elements, and hand structures) that baseline methods frequently fail to reproduce without noticeable artifacts.
(3) Bottom: We provide additional comparative examples between the TPO model and the base model to demonstrate TPO's advantages.

\noindent \textbf{Quantitative Results.}
Tab.~\ref{tab:main} summarizes the quantitative comparisons with baseline methods. While our base model achieves solid lip synchronization (Sync-C = 7.220) and motion richnesss (HKV = 40.022), it falls short of state-of-the-art methods in visual quality, generation fidelity, and hand structure quality. These critical artifacts can make the generated results unusable.
After the proposed timestep-segment preference optimization, the TPO model achieves significant improvements across all evaluation metrics, outperforming all baselines. Both FID and FVD improve by over 10\%, indicating that preference learning enhances overall generation quality and motion naturalness. The TPO model also achieves better hand-related metrics, with an HKC score of 0.910 and an HKV score of 48.604, demonstrating improved structural integrity and motion richness for generated hands. Additionally, the TPO model achieves higher scores in visual quality (IQA: 4.018 → 4.086), aesthetic appeal (AES: 2.876 → 3.002), and lip synchronization accuracy (Sync-C: 7.220 → 7.525).
These results demonstrate the effectiveness of the proposed method, which achieves consistent improvements over strong baselines while balancing  the competing fidelity and motion.

\begin{table}[h]
\caption{Quantitative comparisons with baselines.}
  \centering
  \begin{tabular}{lccccccc}

    \toprule
          Methods &IQA $\uparrow$ & AES$\uparrow$ & Sync-C$\uparrow$ & FID$\downarrow$ & FVD$\downarrow$ &HKV $\uparrow$ & HKC$\uparrow$  \\
    \midrule
    DiffTED~\cite{hogue2024diffted} & 2.701 & 1.703 & 0.926 & 95.455 & 58.871 & - & 0.769 \\
    DiffGest.~\cite{zhu2023taming}+Mimic.~\cite{zhang2024mimicmotion} & 4.041 & 2.897 & 0.496 & 58.953 & 66.785  & 23.409  & 0.833  \\
    CyberHost~\cite{lincyberhost} & 3.990 & 2.884 & 6.627  &  32.972 & 28.003 & 24.733 & 0.884 \\
    OmniHuman~\cite{lin2025omnihuman} & 4.055 &  2.959 & 6.951  & 33.272 &   31.081 & 39.393  &  0.886 \\
    FantasyTalking~\cite{wang2025fantasytalking} & 3.873 & 2.708  & 3.248  & 50.741 & 45.714  &  28.524 & 0.811 \\
    \midrule
    AlignHuman-Base & 4.018 & 2.876 & 7.220 & 33.192 & 32.482 & 40.022
    & 0.869  \\
    AlignHuman-TPO & \textbf{4.086} & \textbf{3.002} & \textbf{7.525} & \textbf{30.004} & \textbf{28.002} & \textbf{48.604} & \textbf{0.910} \\
    \bottomrule
  \end{tabular}
  \label{tab:main}
\end{table}

\subsection{Ablation Study}

\noindent \textbf{Switch timestep. }The switch timestep between motion and fidelity LoRAs is critical, as analyzed previously. We report the TPO model’s performance under different switch timesteps in Tab.~\ref{tab:ablation_t_selection}. During training and inference, motion LoRA is activated before the switch timestep, while fidelity LoRA is activated afterward, enabling the model to optimize distinct aspects at different denoising stages. Switching near 0.2T yields consistently good results since it does not significantly disrupt the learning focus. The best performance is achieved at 0.2T, which aligns with our analysis in Sec.~\ref{sec:analysis}, where the model prioritizes motion optimization in the early denoising stage ($T$ -> $0.8T$) and visual fidelity in the later stage ($0.8T -> 0$). This configuration also corresponds to our final model used for comparison. Performance degrades significantly when the switch timestep deviates substantially from 0.2T. Moreover, switching too early, which compresses the motion learning timestep, leads to a decline in the HKV metric, indicating reduced motion diversity. These observations further validate the rationale behind the timestep-segment preference optimization strategy.
\begin{table}[h]
\caption{Ablation results on switch timestep.}
  \centering
  \resizebox{1.0\textwidth}{!}{
  \begin{tabular}{@{}lccccccc@{}}
    \toprule
    \multicolumn{1}{c}{Methods} & IQA $\uparrow$ & ASE $\uparrow$ & Sync-C $\uparrow$ & FID $\downarrow$ & FVD $\downarrow$ & HKV $\uparrow$ & HKC $\uparrow$ \\
    \midrule

    AlignHuman-TPO-0.10T & 4.084 & 2.984 & 7.405 & 31.943 & 29.915 & 42.048 & 0.909 \\
     AlignHuman-TPO-0.15T & 4.082& 3.000 & 7.441 & 30.965 & 30.053 & 44.187& 0.909 \\
     AlignHuman-TPO-0.18T & 4.083 & 2.995 & 7.515 &30.414 & 28.441 & 47.757 &\textbf{0.910} \\
    AlignHuman-TPO-0.20T & \textbf{4.086} & \textbf{3.002} & \underline{7.525} & \textbf{30.004} & \textbf{28.002} & \textbf{48.604} & \textbf{0.910} \\
     AlignHuman-TPO-0.22T & 4.077 & 2.988 & \textbf{7.534}& 30.288 & 29.230 & 48.428 & 0.897 \\
      AlignHuman-TPO-0.25T & 4.082 & 2.994 & 7.508 & 31.176 & 31.875 & 48.551 & 0.897 \\
     AlignHuman-TPO-0.30T & 4.079 & 2.995 & 7.488 & 33.392 & 33.240 & 47.731 & 0.865 \\
     AlignHuman-TPO-0.40T & 4.081 & 2.998 & 7.512 & 33.858 & 32.978 & 48.335 & 0.852 \\

    \bottomrule
  \end{tabular}
  }
  \label{tab:ablation_t_selection}
\end{table}

\noindent \textbf{Analysis of Key Designs.} In this part, we explore the contribution of each proposed design by conducting several ablation variants, including 1) without Fidelity-LoRA, which optimizes only motion naturalness using preference data; 2) without Motion-LoRA, which optimizes exclusively for visual fidelity preferences; 3) without timestep divide-and-conquer, where preference learning no longer separates motion and fidelity pair data based on timesteps; 4) Single LoRA, which employs a single LoRA module to learn all preferences; 5) Zero LoRA, which performs preference learning by fully fine-tuning the base model; 6) Naive DPO, which averages scores across both dimensions to compute a final preference score for full base model training; and 7) IPO~\cite{azar2024general} and SimPO~\cite{meng2024simpo}, which modify the loss computation in DPO to achieve more stable preference training.

\begin{table}[h]
\caption{Ablation results on key designs.}
  \centering
  \resizebox{1.0\textwidth}{!}{
  \begin{tabular}{@{}lccccccc@{}}
    \toprule
    \multicolumn{1}{c}{Methods} & IQA $\uparrow$ & ASE $\uparrow$ & Sync-C $\uparrow$ & FID $\downarrow$ & FVD $\downarrow$ & HKV $\uparrow$ & HKC $\uparrow$ \\
    \midrule

    w/o Fidelity-LoRA & 4.024 & 2.978 & 7.519 & 32.018 & 30.560 & 46.503 & 0.869 \\
    w/o Motion-LoRA & \textbf{4.086} & 3.001 & 7.517 & 31.853 & 30.620 & 42.695 & \textbf{0.910} \\
    w/o Timestep Segment & 4.081 & 2.998 & \textbf{7.550} & 32.069 & 31.068 & 46.298 & 0.895 \\
    Single Lora & 4.086 & 3.001 & 7.413 & 32.131 & 31.723 & 47.324 & 0.901 \\
    Zero LoRA & 4.046 & 2.969 & 7.429 & 31.574 & 31.663 & 47.324 & 0.890 \\

\midrule
    Naive-DPO & 4.071 & 2.966 & 7.313 & 33.224 & 32.202 & 40.775 & 0.901 \\
    IPO & 4.075 & 2.985 & 7.412 & 32.616 & 31.422 & 41.469 & 0.902 \\
    SimPO & 4.080 & 2.982 & 7.384 & 32.815 & 31.901 & 41.173 & 0.902 \\

    \midrule
    AlignHuman-TPO & \textbf{4.086} & \textbf{3.002} & \underline{7.525} & \textbf{30.004} & \textbf{28.002} & \textbf{48.604} & \textbf{0.910} \\
    \bottomrule
  \end{tabular}
  }
  \label{tab:ablation_key_design}
\end{table}

Tab.~\ref{tab:ablation_key_design} reports the quantitative results. It demonstrates that removing the Fidelity-LoRA leads to significant degradation in both hand quality (HKC$\Delta-0.041$) and visual generation quality (FID$\Delta-2.014$), while removing  the Motion-LoRA results in pronounced degradation in hand motion richness (HKV$\Delta-5.909$) and video quality (FVD$\Delta-2.618$). Removing the timestep segment, while demonstrating improvements over the base model, remains inferior to the full TPO solution. It validates the necessity of decoupling visual fidelity and motion naturalness optimization, as these objectives exhibit competitive relationships. The performance also exhibits degradation when using either a single LoRA or no LoRA modules, confirming the expert LoRA preference modules help decouple the preference optimization for different dimensions. The performance of Naive DPO and other variants achieves comparable performance, moderately improving fidelity but still falling short of the full TPO, while demonstrating negligible motion enhancement. It highlights the necessity of effectively disentangling these two competing dimensions.

\noindent \textbf{Analysis of LoRA Rank. } We further investigate the effects of varying rank sizes in the preference LoRA modules. Scaled from 32 to 512 with factors of 2. Tab.~\ref{tab:lora} shows that the model's performance improves monotonically with increasing the LoRA-rank and achieves saturation around 256.

\begin{table}[h]
\caption{Ablation results on LoRA ranks. The starred (*) represents our final TPO model.}
  \centering
  \scriptsize
  \resizebox{1.0\textwidth}{!}{
  \begin{tabular}{lccccccc}
    \toprule
          Methods &IQA $\uparrow$ & ASE$\uparrow$ & Sync-C$\uparrow$ & FID$\downarrow$ & FVD$\downarrow$ &HKV $\uparrow$ & HKC$\uparrow$  \\
    \midrule
    rank 32 & 4.078 & 3.001 & 7.401 & 32.558 & 30.551  & 44.956  & 0.895  \\
    rank 64 & 4.075 & 2.999 & 7.476  &  31.365 & 30.158 & 45.498 & 0.905 \\
    rank 128 & 4.085 &  3.003 &  7.516  & 30.321 &   28.223 & 44.404  &  0.908 \\
    rank 256$^*$  & \textbf{4.086} & 3.002 & \textbf{7.525} & \textbf{30.004} & \textbf{28.002} & \textbf{48.604} & \textbf{0.910}  \\
    rank 512 & \textbf{4.086} & \textbf{3.005} & 7.521 & 30.017 & 28.014 & 48.392
    & \textbf{0.910}  \\
  \bottomrule
  \end{tabular}
  }
  \label{tab:lora}
\end{table}

\subsection{Discussions on Accerating}
\begin{figure}[htbp]
    \centering
    \begin{subfigure}[b]{0.32\textwidth}
        \includegraphics[width=\linewidth]{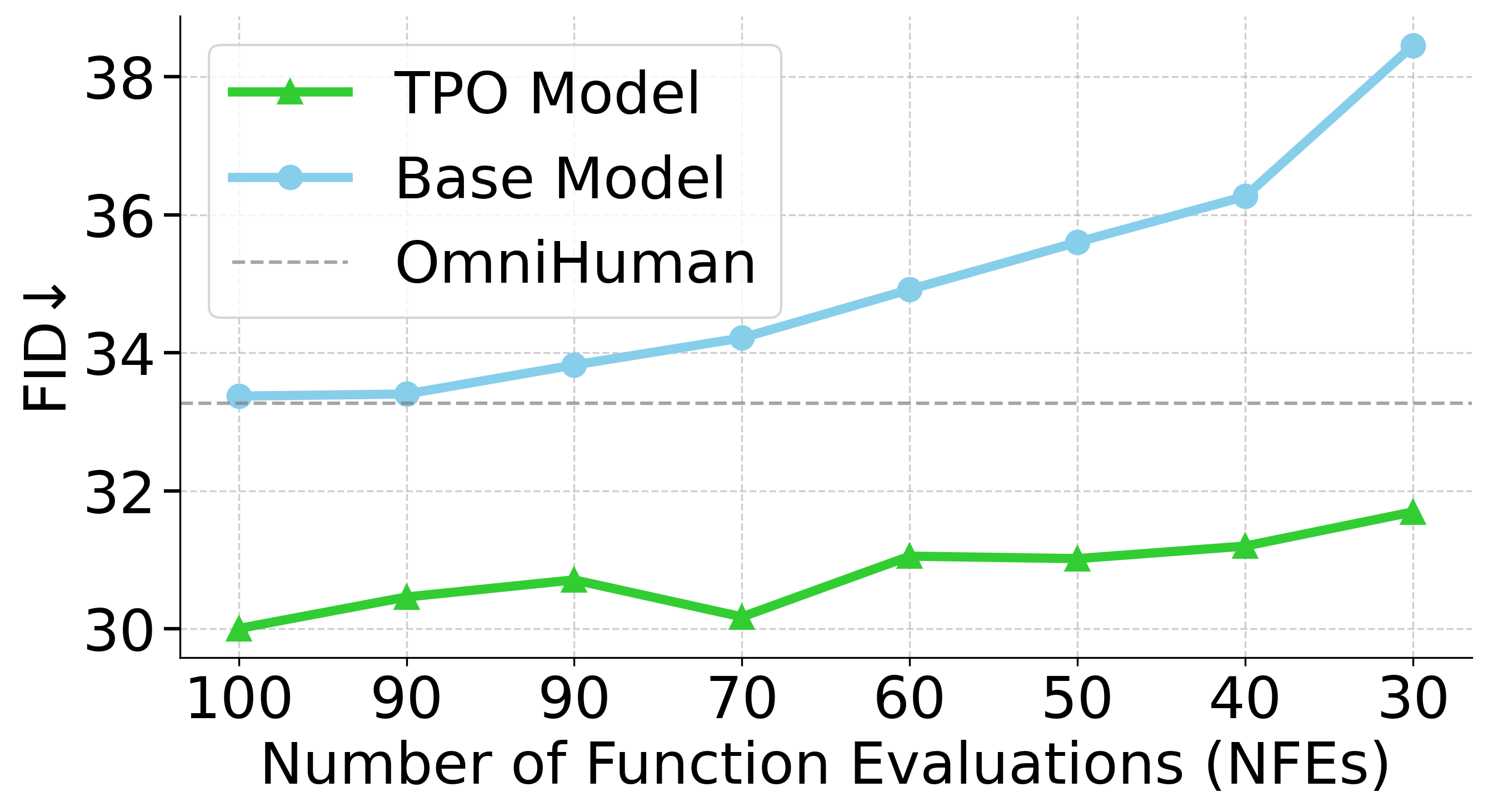}
        \caption{NFEs-FID}
        \label{fig:accelerate_FID}
    \end{subfigure}
    \hfill 
    \begin{subfigure}[b]{0.32\textwidth}
        \includegraphics[width=\linewidth]{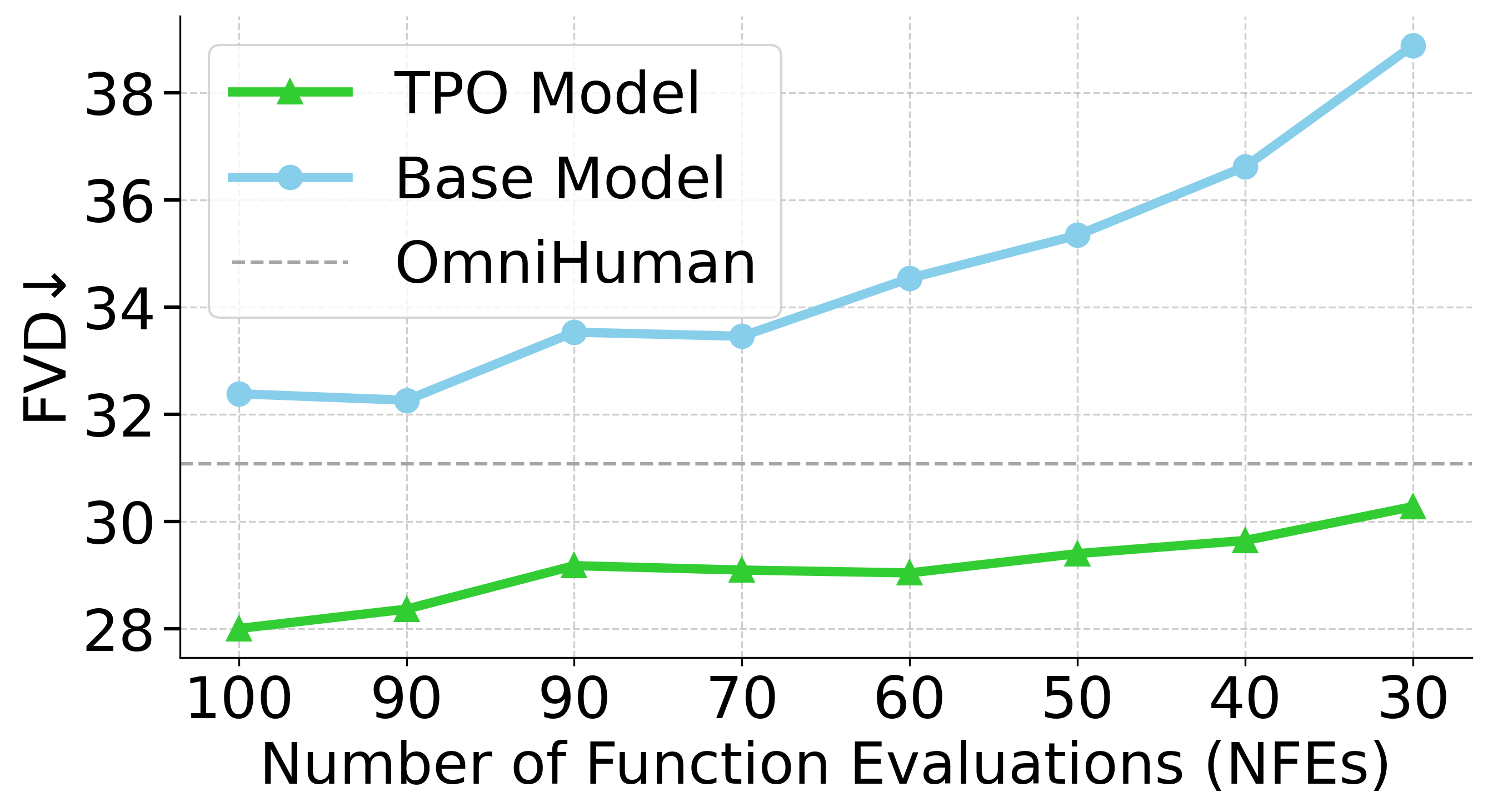}
        \caption{NFEs-FVD}
        \label{fig:accelarate_FVD}
    \end{subfigure}
    \hfill
    \begin{subfigure}[b]{0.32\textwidth}
        \includegraphics[width=\linewidth]{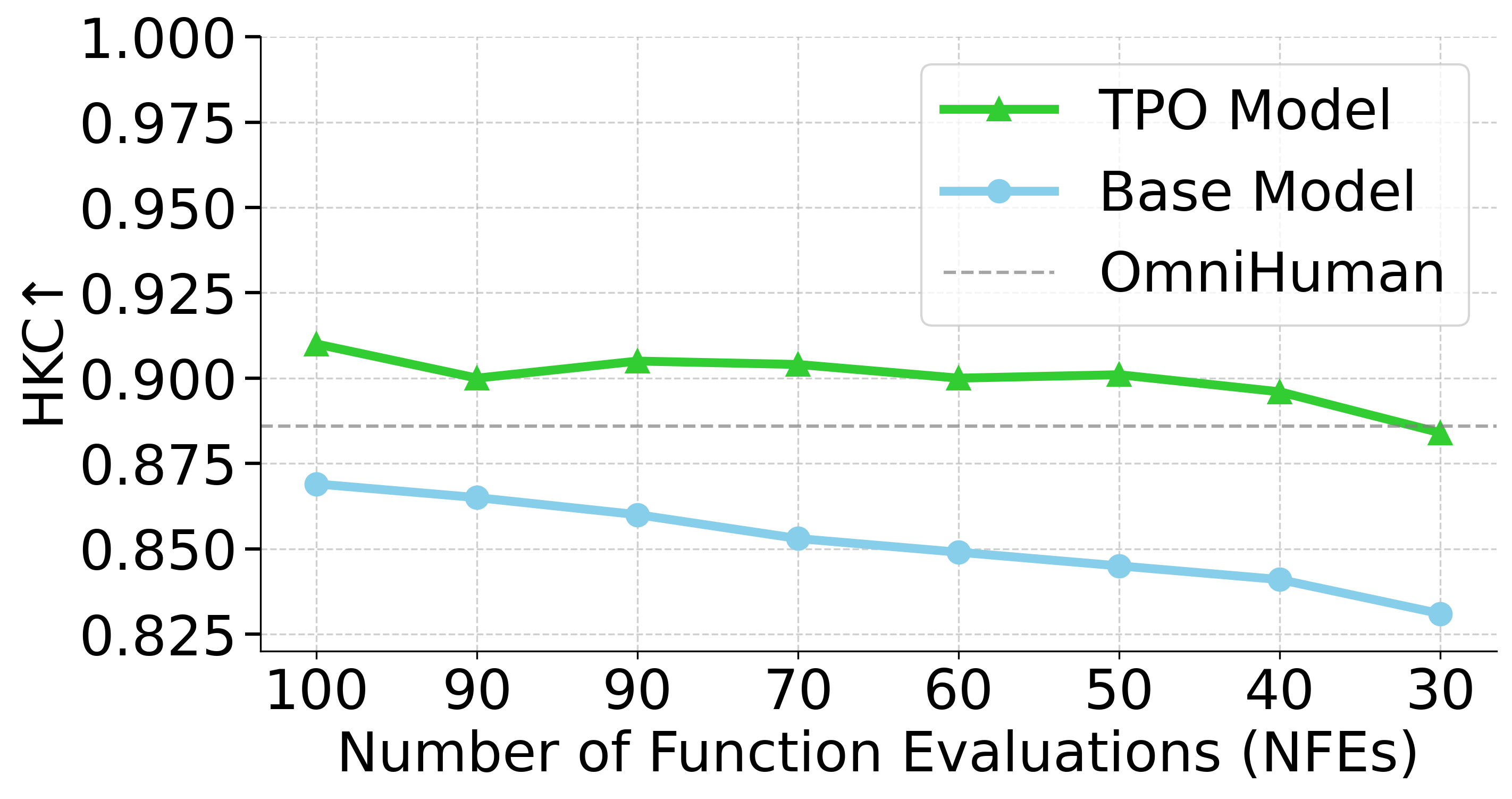}
        \caption{NFEs-HKC}
        \label{fig:accelerate_HKC}
    \end{subfigure}
    \caption{Performance comparison of the TPO model and the base model under different NFEs.}
    \label{fig:accelerate}
\end{figure}

\begin{figure}[htbp]
    \centering
    \includegraphics[width=\linewidth]{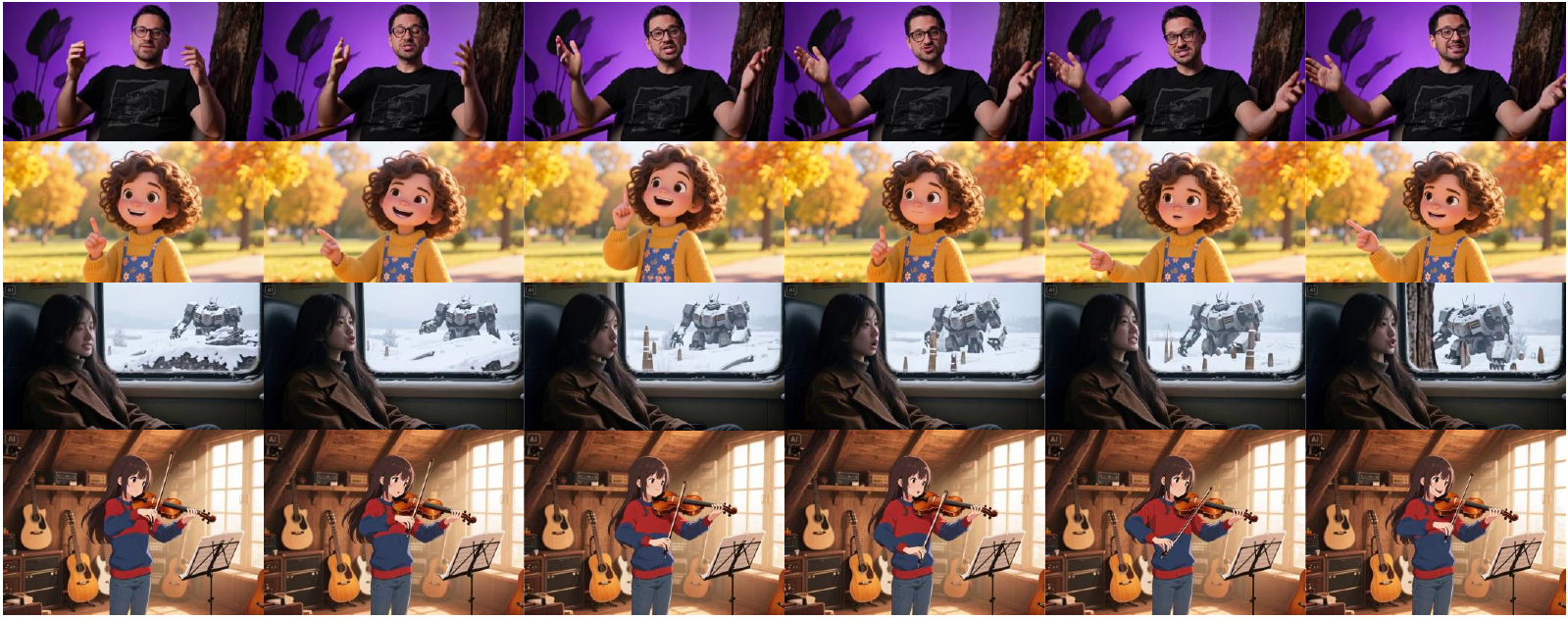}
        \caption{Visualization results of our TPO model at 30NFEs.} 
        \label{fig:vis_20steps}
\end{figure}

The timestep-segment preference optimization disentangles the motion and visual dimensions, enabling the TPO model to surpass both the base model and state-of-the-art baselines. Here, we further investigate the impact of the TPO model on inference steps. Starting from 100 NFEs (50 denoising steps with one CFG calculation per step), we progressively reduce NFEs by 10 during inference and evaluate performance. Fig.~\ref{fig:accelerate} shows the performance trends of the TPO model compared to the base model across metrics including HKC, FID, and FVD, alongside the benchmark results of the state-of-the-art OmniHuman-1\cite{lin2025omnihuman} model (100 NFEs). Notably, even with only 30 NFEs, the TPO model significantly outperforms the base model at 100 NFEs and achieves comparable results to OmniHuman. Visualizations of the TPO model with 30 NFEs are presented in Fig.~\ref{fig:vis_20steps}. Both quantitative and qualitative results demonstrate the effectiveness of timestep-segment preference learning in stabilizing the denoising process, enabling the generation of high-quality intermediate latents and thus achieving inference speedup.

\section{Conclusion}
In this work, we address the challenge of balancing motion naturalness and visual fidelity in audio-driven human animation by proposing AlignHuman, a novel framework that incorporates timestep-segment preference optimization (TPO) with a divide-and-conquer strategy. Through an in-depth analysis of the denoising process, we uncover that motion dynamics are primarily determined in early denoising timesteps, while fidelity benefits from later stages. Leveraging this insight, AlignHuman employs specialized LoRA modules for motion and fidelity, activated at specific timestep intervals during training and inference. This design effectively mitigates the competing relationship between motion and fidelity, enabling the model to optimize both objectives without compromise. Extensive experiments demonstrate that AlignHuman not only achieves state-of-the-art performance but also significantly reduces inference NFEs, delivering a 3.3× speed-up with minimal impact on generation quality. Our work highlights the importance of timestep-segment optimization in diffusion-based models and provides a robust solution for expressive and realistic human animation.

\clearpage
{
    \small
    \bibliographystyle{plain}
    \bibliography{main}
}



\end{document}